\documentclass[11pt]{article}

\usepackage[preprint]{neurips_2021}

\usepackage[utf8]{inputenc} 
\usepackage[T1]{fontenc}    
\usepackage{hyperref}       
\usepackage{url}            
\usepackage{booktabs}       
\usepackage{amsfonts}       
\usepackage{nicefrac}       
\usepackage{microtype}      
\usepackage[dvipsnames,table,xcdraw]{xcolor}
\usepackage{graphicx}
\usepackage{amsmath}
\usepackage{multirow,array}
\usepackage{algorithm}
\usepackage{algpseudocode}
\newcommand{\floor}[1]{\lfloor #1 \rfloor}

\title{A Short Study on Compressing Decoder-Based Language Models}

\author{Tianda Li$^*$, Yassir El Mesbahi\thanks{Equal contribution}  , Ivan Kobyzev, Ahmad Rashid,\\
\textbf{Atif Mahmud, Nithin Anchuri, Habib Hajimolahoseini, Yang Liu,}\\ 
\textbf{Mehdi Rezagholizadeh} \\
Huawei Noah’s Ark Lab\\
  {\texttt \{tianda.li, yassir.el.mesbahi1,ivan.kobyzev, ahmad.rashid, mehdi.rezagholizadeh\}@huawei.com}\\
}

\usepackage{ulem}
\usepackage{color}

\begin{document}
\colorlet{shadecolor}{gray!25}
\maketitle

\begin{abstract}
Pre-trained Language Models (PLMs) have been successful for a wide range of natural language processing (NLP) tasks. The state-of-the-art  of PLMs, however, are extremely large to be used on edge devices. As a result, the topic of model compression has attracted increasing attention in the NLP community. 
Most of the existing works focus on compressing encoder-based models (tiny-BERT, distilBERT, distilRoBERTa, etc), however, to the best of our knowledge, the compression of decoder-based models (such as GPT-2) has not been investigated much. Our paper aims to fill this gap. Specifically, we explore two directions: 1) we employ current state-of-the-art knowledge distillation techniques to improve  fine-tuning of DistilGPT-2. 
2) we pre-train a compressed GPT-2 model using layer truncation and compare it against the distillation-based method (DistilGPT2). The training time of our compressed model is significantly less than DistilGPT-2, but it can achieve better performance when fine-tuned on downstream tasks. We also demonstrate the impact of data cleaning on model performance.

\end{abstract}

\section{Introduction}
Pre-trained Language Models (PLMs) have recently achieved great success on a wide variety of NLP problems~\citep{Peters:2018,devlin2019bert,liu2019roberta,yang2020xlnet,Radford2018ImprovingLU,radford2019language}.
With the rapidly increasing parameter count and training time, the state-of-the-art (SOTA) PLMs are becoming more challenging to be deployed on edge devices. In particular, RoBERTa-large has 355 million parameters, GPT-2-xl has 1.5 billion parameters, and the most recent GPT-3~\citep{brown2020language} has 175 billion parameters. The importance of model compression methods is emergent in NLP~\citep{gupta21_survey}. 

Generally speaking, the compression of a PLMs can be divided into two stages: initialization and fine-tuning. In the initialization stage, the compressed model's parameters can be either transferred from a larger pre-trained model~\citep{sun2019patient,passban2020alpkd} or pre-trained from scratch as a language model. Pre-training the smaller language models is cumbersome since typically knowledge is distilled from a larger teacher~\citep{jiao-etal-2020-tinybert,sanh2020distilbert}. 
In the fine-tuning stage, the initialized compressed model is trained on a downstream task. In our work, we will investigate both stages of the compression. 



\begin{figure}[tb]
    \centering
    \includegraphics[width=14cm,height=7cm]{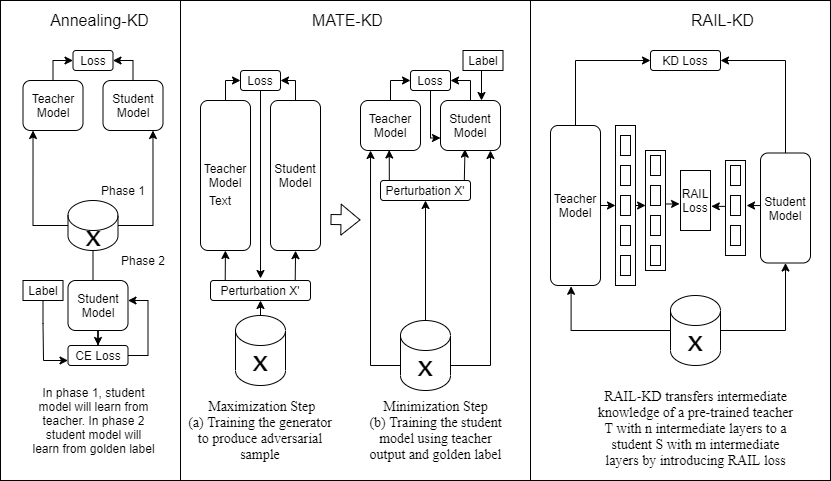}
    \caption{The overview of knowledge distillation techniques we apply in the paper: Annealing-KD, MATE-KD and RAIL-KD's.}
    \label{fig:model}
\end{figure}

A predominant solution for fine-tuning compressed PLMs is knowledge distillation (KD)~\citep{rogers-etal-2020-primer}.
Most of the reported KD results in the literature~\citep{hinton2015distilling,bucilua2006model,8424632,kamalloo-etal-2021-far,rashid2020towards,ti2021select,haidar2021railkd} are for encoder-based models such as BERT, RoBERTa. KD on decoder-based models~\citep{radford2019language} has not been investigated much. In this work we explore both teacher-based and teacher-free techniques aiming to improve compressed GPT-2 fine-tuning.

Pre-training of compressed encoder-based models has been extensively explored \citep{sanh2020distilbert, Xu2020BERTofTheseusCB, poormansbert}. 
However, DistilGPT2~\citep{distilgpt2} is the only compressed GPT-2 model we found in the literature. The authors pre-trained DistilGPT-2 with KD using the original GPT-2 as a teacher, which results in a long training time. In our paper, we investigate pre-training without KD to significantly improve time efficiency. 

Our contribution is three-fold:
\begin{enumerate}
 \item We benchmark different SOTA teacher-based and teacher-free techniques for fine-tuning DistilGPT2 on downstream tasks. 
    \item We compare several truncation methods for pre-training initialization of the compressed GPT-2 model.
    \item We conduct data cleaning of the OpenWebText dataset and pre-train our compressed model initialized with the best truncation technique. This pre-training scheme is time-efficient. At the same time, fine-tuning on downstream tasks reveal that our pre-trained model achieves better performance compared to DistilGPT-2.
\end{enumerate}

\section{Methodology}

In this section, we introduce the techniques we applied for fine-tuning and pre-training compressed GPT-2 models. We start with knowledge distillation and teacher-free methods for fine-tuning, then we introduce layer truncation methods for initializing the student from the teacher, and finally, we discuss data-cleaning for efficient pre-training.

\subsection{Fine-tuning with and without a teacher}
\label{s:kdtf}
Here, we discuss the techniques we applied to improve fine-tuning of DistilGPT-2 model on downstream tasks. 
\subsubsection{KD Methods}
\citet{hinton2015distilling} proposed KD as a way to improve the training of a small neural network (student). Given a bigger model (teacher), KD adds a specific loss term to the loss function of the student aiming to push the student's predictions close to the teacher's.
In this paper, we consider four different KD methods: 1) Vanilla KD, 2) Annealing-KD, 3) MATE-KD and 4) RAIL-KD. The overview of these models is given in Figure~\ref{fig:model},

For \textbf{Annealing-KD}~\citep{Aref2021}, the student is trained in two phases. During phase 1, the student model learns only from the teacher. Here the temperature controls the smoothness of the teacher's output, annealing it from easy-to-learn to the actual sharp distribution. During phase 2, the student model is trained only on the ground-truth label. 

For \textbf{MATE-KD}~\citep{rashid-etal-2021-mate}, the training process has two steps: maximization and minimization. At the maximization step, a generator is trained to produce perturbed input for both student and teacher models. The target of this stage is to produce the input that can maximize the divergence between the teacher and student output. At the minimization step, the student model is trained to approximate the teacher's output.

For \textbf{RAIL-KD}~\citep{haidar2021railkd}, 
during the training we transfer the knowledge from teacher's intermediate layers to student's  intermediate layers. In our case, the 6-layers student model is distilled from 12 layers GPT2 model.


\subsubsection{Teacher-free Methods}
Here, we describe the most commonly used teacher-free techniques.  

\paragraph{Label Smoothing (LS)}~\citet{DBLP:journals/corr/SzegedyVISW15} proposed this method to improve the training of a classifier. 
For this, a cross-entropy loss should be calculated with smoothed labels rather than one-hot labels. The smoothed labels are given by:
    \begin{equation}
\label{eq:ls}
    y' = (1-\alpha) y + \alpha u ,
\end{equation}
where $u(K) = 1/K$ is the uniform distribution on $K$ classes, $y$ is the one-hot golden label, and $\alpha$ is a parameter between 0 and 1 controlling the sharpness of the resulting soft label.
\paragraph{TF-reg} The TF-reg~\citep{yun2020} technique is very similar to label smoothing, the only difference is that TF-reg switches the uniform distribution $u$ in Equation \ref{eq:ls} to the label-dependent distribution $p(k)$, defined by:
\begin{equation}
\label{eq:p_tfreg}
    p^c(k) = \begin{cases} a, \ \text{if} \  k=c \ \text{(is the correct label)}  \\ \frac{1-a}{K-1}, \ \text{otherwise.} \end{cases}
\end{equation}
Where $a$ is a parameter between 0 and 1. 
TF-reg has two parameters ($a$ and $\alpha$) instead of just one ($\alpha$) which allows for better tuning. The smoothed label for $x$ in TF-reg is given by:
\begin{equation}
\label{eq:tfreg}
    y' = (1-\alpha) y + \alpha  p^{c(x)} ,
\end{equation}
where $c(x)$ is the correct label of the sample $x$.

\paragraph{Self-distillation (Self-KD)} Self-KD~\citep{furlanello2018born} is a  variation of the KD method, in which we first fine-tune a copy of the student on the dataset and then freeze it. This copy  serves as a teacher during the training.

\subsection{Student Layers Initialization}
\label{sec-init}
In this section, we introduce the student's layers initialization from the teacher.
\citet{poormansbert} shows that an easy way of compressing pre-trained large models is to simply "truncate" them by dropping some layers. Inspired by that, we propose our pruning techniques and list the top 2 pruning strategies below (The overall six pruning techniques and results are introduced in Appendix A.1) 

\paragraph{Uniform Selection (Uni)} We select layers to copy uniformly, starting from the first layer. For example, if the teacher has $12$ layers, we would initialize the student (6-layers model) by teacher layers $0$, $2$, $4$, $6$, $8$ and $10$. 
\paragraph{Pseudo-uniform Selection (Psudo)} This strategy is inspired from DistilBert's paper \citep{sanh2020distilbert}, where they initilaize their model (DistilBert) with teacher's (Bert-base) layers $0$, $2$, $4$, $7$, $9$ and $11$. In contrast with uniform selection, we make sure first and last layers are always selected.
A generalization of this strategy can be described by the Algorithm \ref{alg:pseud-uniform}, where $n$ stands for the total number of teacher's layers and $k$ is number of layers we want to select (also number of student's layers).

\begin{algorithm}[tb]
\caption{Details of Pseudo-uniform selection for layer initialization from a larger model with $n$ layers to a smaller model with $k$ layers.}
\label{alg:pseud-uniform}
\begin{algorithmic}
\Require $n > k$; $n\,\,\text{mod}\,k = 0$; $n\,\,\text{mod}\,2 = 0$
\Ensure Pseudo-uniform selection of length $k$
\State step $\gets \floor{\frac{n}{k}}$
\State start $\gets 0$
\State end $\gets n-1$
\State selection $\gets []$
\While {start $\leq$ end}
\State selection $\gets $ selection $+ [\text{start}]$ 
\State selection $\gets $ selection $+ [\text{end}]$ 
\State start $\gets \text{start} + \text{step}$
\State end $\gets \text{end} - \text{step}$
\EndWhile
\end{algorithmic}
\end{algorithm}


\section{Experiments}

\subsection{Data}
OpenWebText is an open-source recreation of the WebText corpus (on which the GPT-2 model was trained). We use this data for pre-training our compressed model. Original WebText contains over 8 million documents for a total of 40 GB of text. In our experiment, we only used a fraction of these data. 

We assess compressed models on several downstream tasks. 
First, we employ the Wikitest103 dataset~\citep{DBLP:journals/corr/MerityXBS16} to fine-tune a compressed model as a language model and measure the performance with perplexity score (the lower - the better). 
Then, we fine-tune a compressed model as a classifier  on 6 out of 8 tasks in the SuperGLUE~\citep{DBLP:journals/corr/abs-1905-00537} benchmark. 
Moreover, we evaluate the fine-tuning of a compressed model as a classifier on 7 out of 9 tasks of the General Language Understanding Evaluation
(GLUE)~\citep{wang2019glue} benchmark.

\subsection{Fine-tuning on GLUE}
We apply KD and teacher-free techniques described in Section~\ref{s:kdtf} to fine-tune DistilGPT-2 model on GLUE tasks. The results are in Table~\ref{tab:teacher_free} and Table~\ref{tab:ComKD_glue}.  
We can see that Annealing-KD, MATE-KD, and RAIL-KD all outperform  VanillaKD. 
Interestingly, regular fine-tuning itself is a strong baseline that performs comparatively well to vanilla KD, and it even outperforms the LS and TF-reg techniques. Self-KD performance is comparable with other teacher-free techniques.
RAIL-KD performs worse than MATE-KD and Annealing-KD, which indicates that distilling intermediate layers doesn't have an advantage over data augmentation or annealing scheduling. 
MATE-KD performs the best among four KD techniques. One should notice that this pattern is slightly different from the fine-tuning of Bert-based models~\citep{ti2021select}. 
One possible explanation might be that decoder-based models are more sensitive to hyper-parameters. Data augmentation is a more robust way to improve the student model's performance.


\begin{table}[!htb]
 \caption{Dev set results of teacher-free methods on GLUE. We benchmark pure finetuning of DistilGPT2 (first line) with teacher-free regularisation training and Self-KD. The last line is the performance of 12 layers GPT-2 model.}
 \label{tab:teacher_free}
\small
\begin{center}
\resizebox{\columnwidth}{!}{	
\begin{tabular}{c|c|c|c|c|c|c|c|c} 
    \hline
    \toprule
     Evaluated Model & CoLA & RTE & MRPC(f1) & SST-2 & MNLI & QNLI & QQP & Average \\
    \midrule
    DistilGPT2 & 39.0  & 65.2 & 87.9  & 91.5   & 86.5  & 79.9  & 89.7  & 77.1  \\[0.35ex]
    LS &  38.9   & 64.8   & 87.3   & 91.6  & 86.6 & 80.1   & 89.6    & 77.0   \\
    TF-reg & 38.7   & 65.1  & 87.4   & 91.4   & 86.9   & 80.2    & 89.6   & 77.0    \\
    Self-KD &  39.7   & 64.7   & 87.3   & 90.9   & 87.0   & 80.5   & 89.8 & 77.2   \\ 
    GPT-2    & 43.2    & 66.8    & 87.6    & 92.2    & 82.3    & 88.6    & 89.5    & 78.6 \\[0.35ex]
    \bottomrule
\end{tabular}}
\end{center}
\end{table}

\begin{table}[!htb]
 \caption{Dev set results of KD methods on GLUE. Here the student is DistilGPT2 and the teacher is 12 layers GPT-2.  See Table~\ref{tab:teacher_free} for the student's and teacher's performance.
}
 \label{tab:ComKD_glue}
\small
\begin{center}
\resizebox{\columnwidth}{!}{	
\begin{tabular}{c|c|c|c|c|c|c|c|c|c} 
    \hline
    \toprule
     Teacher & Evaluated Model & CoLA & RTE & MRPC(f1) & SST-2 & MNLI & QNLI & QQP & Average \\
     \midrule
    GPT-2    & $\text{VanillaKD}_{\text{DistilGPT2}} $   & 39.3    & 65.7    & 88.0    & 90.7    & 79.6    & 86.8    & 89.4    & 77.1 \\[0.35ex]
    GPT-2    & $\text{RailKD}_{\text{DistilGPT2}}$    & 39.4    & 66.4    & 88.1    & 91.2    & 80.6    & 87.3    & 89.9    & 77.6  \\[0.35ex]
    GPT-2    & $\text{AnnealingKD}_{\text{DistilGPT2}}$    & 41.6    & 67.1    & 86.8    & 92.0    & 80.8    & \textbf{87.8}    & 89.4    & 77.9   \\[0.35ex]
    GPT-2    & $\text{MateKD}_{\text{DistilGPT2}}$    & \textbf{42.1}    & \textbf{67.5}    & \textbf{88.8}    & \textbf{92.0} & \textbf{81.6}  & 87.7 & \textbf{90.0}  & \textbf{78.5}   \\[0.35ex]
    \bottomrule
\end{tabular}}
\end{center}
\end{table}

\subsection{Experiments on Layer Truncation}
First, we initialize a 6-layer GPT-2 model with the initialization techniques described in section \ref{sec-init}.
Then, we pre-train  the models on fraction of the OpenWebText dataset. 
For these experiments, we use either 4 or 8 GPUs and make use of the DeepSpeed framework \citep{deepspeed1, deepspeed2, deepspeed3, deepspeed4} to accelerate the training process. Then, we report the zero-shot performance of the pre-trained models in Table~\ref{tab:trun-pretrain}. Our compressed model outperforms DistilGPT-2 even when it's trained on $50\%$ of the dataset. Also, our pre-training is  tremendously  time-efficient.
\begin{table}[!htb]
\caption{Truncated models' zero-shot perplexity scores on Wikitext103 after pretraining. Models are truncated with techniques from Section~\ref{sec-init}, pre-trained on fraction of the OpenWebtext dataset, and then evaluated on Wikitext103 test set. All the truncated models in this table have 6 layers.}
\label{tab:trun-pretrain}
\small
\begin{center}
\begin{tabular}{c|c|c|c|c|c|c|c} 
    \hline
    \toprule
    \multicolumn{2}{c|}{Models} & \multicolumn{4}{c}{Pretraining on fraction of OpenWebtext}\\
    \cmidrule{0-6}
      \multicolumn{1}{c|}{Model index}&\multicolumn{1}{c|}{Teacher} &\multicolumn{1}{c|}{Strategy}& \multicolumn{1}{c|}{PPL} & \multicolumn{1}{c|}{$\%$ of the dataset}&\multicolumn{1}{c|}{Epochs}&\multicolumn{1}{c|}{\# GPUs}&\multicolumn{1}{c}{Time (h)}\\
    \midrule
    \rowcolor{shadecolor}
    0&DistilGPT2& Pre-train  & \bf 45.26 &  100  & 4 & 8  & 768 \\
    \midrule
    1&GPT2& Psudo& 56.38 & 10 & \multirow{5}{*}{3} & 4 &  8 \\[0.35ex]
    2&GPT2& Psudo & 46.91  & 50 &  & 8 &  38 \\[0.35ex]
    3&GPT2& Uni  & \bf 45.19   &  50 &  & 8 &  35 \\[0.35ex]
    4&GPT2-xl& Psudo & 54.70  & 10 & & 4 & 24 \\[0.35ex]
    5&GPT2-large& Psudo & 59.32 & 10 &  & 4 & 17 \\[0.35ex]
    \bottomrule
\end{tabular}
\end{center}
\end{table}

Next, we fine-tuned the pre-trained models on the Wikitext103 and put the perplexities in Table~\ref{tab:trun-pretrain-finetune}. 
We can see that the perplexity achieved by GPT2-psudo is still worse than DistilGPT2's. The 6-layer model truncated from GPT2-xl teacher and initialized with Pseudo-uniform truncation method (GPT2-xl-psudo) reaches a perplexity close to DistilGPT2's despite being pre-trained on a fraction ($10\%$) of the OpenWebtext dataset (but it is not comparable to DistilGPT2 since it has three times more parameters).
\begin{table}[!htb]
\caption{Truncated models' perplexity scores on Wikitext103 after fine-tuning. Once models are pre-trained on fractions of OpenWebText (table \ref{tab:trun-pretrain}), they are fine-tuned on Wikitext103 train set and then evaluated on Wikitext103 test set. All the models compared in this table have 6 layers and the model index indicates corresponds the one in table~\ref{tab:trun-pretrain}.}
\label{tab:trun-pretrain-finetune}
\small
\begin{center}
\begin{tabular}{c|c|c|c|c|c|c|c} 
    \hline
    \toprule
    \multicolumn{2}{c|}{Models} & \multicolumn{4}{c}{Fine-tuning pretrained models}\\
    \cmidrule{0-5}
     \multicolumn{1}{c|}{Model index}&\multicolumn{1}{c|}{Teacher} &\multicolumn{1}{c|}{Strategy}&  \multicolumn{1}{c|}{PPL} & \multicolumn{1}{c|}{Epochs}&\multicolumn{1}{c|}{\# GPUs}&\multicolumn{1}{c}{Time (h)}\\
    \midrule
    \rowcolor{shadecolor}
    0&DistilGPT2& Pre-train  & \bf 21.13 &  6 & 4  & 2 \\
    \midrule
    1&GPT2& Psudo & 23.44 &  12 & 4 &  5 \\[0.35ex]
    2&GPT2& Psudo & 22.67  & 6  & 8 &  2 \\[0.35ex]
    3&GPT2& Uni & 22.61 & 6 & 4 &  5 \\[0.35ex]
    4&GPT2-xl& Psudo & \bf 21.30  & 6 & 4 & 5 \\[0.35ex]
    5&GPT2-large& Psudo & 23.44 & 6 & 4 & 5 \\[0.35ex]
    \bottomrule
\end{tabular}
\end{center}
\end{table}
\subsection{Effect of Data Cleaning on Pre-training}
We found that the OpenWebText dataset contains a significant amount of noisy samples. Some of these are HTML code, others are pure noise (concatenation of special characters). To alleviate the problem of noisy samples, we implemented a program that automatically inspects the samples, clears out HTML code and short sentences, eliminates sentences with a high ratio of non-alphanumerical characters (more than $10\%$) and duplicates.
Using the above algorithm, we managed to dramatically reduce the size of the OpenWebText dataset (from $332,011,430$ to $114,366,559$ samples, or by $65.5 \%$).

We  pre-train a 6-layer GPT2 model (initialized with the pseudo-uniform strategy) on the cleaned dataset, then we fine-tune it on the Wikitext103 and several datasets from the SuperGLUE and compare the results with the model that has been pre-trained on the full dataset. For Wikitext103, we measure zero-shot (ZS) and post-fine-tuning (FT) perplexities (PPL). Results are shown in Table~\ref{tab:dataset-cleaning}.

\begin{table}[!htb]
\caption{Results after pre-training on regular/cleaned OpenWebText dataset. Truncated models are pretrained on both datasets (original and cleaned) and their performance measured on several tasks.}
\label{tab:dataset-cleaning}
\begin{center}
\resizebox{\columnwidth}{!}{	
\begin{tabular}{c|c|c|c|c||c||c||c|c||c||c||c|c} 
    \hline
    \toprule
    \multirow{3}{*}{Models} & \multirow{3}{*}{Pretrain} & \multirow{3}{*}{Pretrain} &\multicolumn{10}{c}{Scores after fine-tuning}\\
    \cmidrule{4-13}
     &  \multirow{2}{*}{dataset} &  \multirow{2}{*}{time (h)} & \multicolumn{2}{c||}{Wikitext103}& \multicolumn{1}{c ||}{BoolQ} & \multicolumn{1}{c ||}{Copa} &\multicolumn{2}{c||}{CB} & \multicolumn{1}{c ||}{Rte} & \multicolumn{1}{c||}{Wic} & \multicolumn{2}{c}{Wsc}\\
     \cmidrule{4-13}
      &  &  & PPL (ZS) &  PPL (FT) & Acc. & Acc. & Acc. &  F1 & Acc. & Acc. &  F1 & Acc. \\
    \midrule
    \rowcolor{shadecolor}
    DistilGPT2 & Regular &  768  & 45.26 & 21.13 & 71.16 & 56 & 73.21 & 61.45 & 62.45 & 63.6 & 63.46 & 77.64\\
    \midrule
    \multirow{2}{*}{GPT2-Psudo}  & Cleaned &  \textbf{42} & 45.93 & 22.42 &  \textbf{70.67} & \textbf{58} & \textbf{78.57} & \textbf{65.15} & \textbf{63.53} & \textbf{60.3} & \textbf{63.46} & \textbf{77.64}\\[0.15ex]
     & Regular &  49 & \textbf{44.51} & \textbf{22.29} & 68.07 & 56 & 71.42 & 49.77 & 58.48 & 59.24 & 59.61 & 73.74\\
    \bottomrule
\end{tabular}}
\end{center}
\end{table}

We can see that cleaning the dataset helps reducing training time while allowing for achieving comparable or better performance. 

\section{Conclusion}

    In this work, we aim to compress GPT-2.
    First, we benchmark current SOTA KD and teacher-free methods on DistilGPT2 and pick the best performing one.
    Then, we explore truncation methods for the initialization of the student model from the teacher model's parameters. Specifically, we propose a pseudo-uniform strategy that outperforms alternative initializations in the language modeling experiments on Wikitext-103. Finally, we conduct data cleaning on the OpenWebText dataset and pre-trained our compressed model. To test the effectiveness of our strategy we carried out the experiments on Wikitext-103 and 6 out of 8 SuperGLUE Benchmark datasets. Our pre-trained model outperforms DistilGPT-2 on 5 out of 7 downstream tasks, yet it is significantly more time-efficient. For the future direction, we will evaluate our initialization strategy along with KD methods and investigate if the pre-training of our compressed GPT-2 can be improved even more.

\normalem
\bibliographystyle{abbrvnat}
\bibliography{nips2021.bib}

\newpage

\appendix

\section{Appendix}

\subsection{Student Layer Initialization}
\label{Layer_Initialization}

Overall, we tried 6 pruning strategies as listed below:
\begin{description}
\item [Uniform selection] We select layers to copy uniformly, starting from the first layer. For example, if the teacher has $12$ layers, we would initialize the student (6-layers model) by teacher layers $0$, $2$, $4$, $6$, $8$ and $10$. 
\item [Variant of uniform selection] We select layers to copy uniformly, but the last layer we select should always be the one before the last of the teacher's layers.
\item [Pseudo-uniform selection] This strategy is inspired from DistilBert's paper, where they initilaize their model(DistilBert) with teacher's (Bert-Base) layers $0$, $2$, $4$, $7$, $9$ and $11$. In contrast with uniform selection, we make sure first and last layer are always selected. 
where $n$ represents the total number of teacher's layers and $k$ the number of layers we want to select (also number of student's layers).
\item [Bottom half selection]
This is a generalization of one of the strategies describe in their paper. We uniformly select from the bottom-half section of teacher's layers. As an example, for a 36-layer teacher, we select uniformly from its first 13 layers. In the particular case of a 12-layer teacher, we pick the first six layers.
\item [Top half selection]
Similar to the bottom half selection, this strategy consists in selecting layers uniformly from the top layers of the teacher. For example, for a 48-layer teacher, we would select uniformly from its top 24 layers. In the particular case of a 12-layer teacher, we pick the last six layers.
\item [Random selection]
We implement this method to have a baseline to compare with. We randomly pick layers from the teacher, sort them by index and use them to initialize the student.
\end{description}

We apply the above pruning techniques on the GPT-2 models and measure their perplexities on Wikitext103 after fine-tuning. To easily identify the models we are training, we add to the names of original GPT-2 a suffix indicating which layer selection strategy was used to initialize it. Table \ref{tab:suffixes} shows the correspondence between suffixes and pruning strategies. Table \ref{tab:trun-finetune} displays some characteristics of the resulting models, as well as their performance on Wikitext103 test set.
\begin{table}[!htb]
\caption{Pruning strategies suffixes}
\small
\begin{center}
\begin{tabular}{c|c} 
    \hline
    \toprule
    Strategy & Suffix \\
    \midrule
    Uniform & uniform\\[0.35ex]
    Uniform (variant) & uniform-2 \\[0.35ex]
    Pseudo-uniform &  psudo \\[0.35ex]
    Bottom-half &  6bh\\[0.35ex]
    Top-half &  6th\\[0.35ex]
    Random & random  \\[0.35ex]
    \bottomrule
\end{tabular}
\end{center}
\label{tab:suffixes}
\end{table}
\begin{table}[!htb]
\caption{Truncated models' perplexity scores on Wikitext103 test set}
\tiny
\begin{center}
\begin{tabular}{c|c|c|c|c|c|c|c} 
    \hline
    \toprule
    Models & \# layers & Teacher's layers & \# heads & Hidden size & \# parameters &  Epochs & PPL \\
    \midrule
    \rowcolor{shadecolor}
    DistilGPT2 & 6 & $\frac{~~~~~~~~~~~~}{~~~~~~~~~~~~}$ & 12& 768 & 81 M & 6 & \bf 21.13 \\
    \midrule
    GPT2-xl-psudo &\multirow{6}{*}{6} & $\{0, 8, 16, 31, 39, 47\}$ & \multirow{6}{*}{25}& \multirow{6}{*}{1,600} & \multirow{6}{*}{266 M} & \multirow{6}{*}{6} &  \bf 22.8\\[0.35ex]
    GPT2-xl-uniform &  & $\{0, 9, 18, 27, 36, 45\}$ &  &  &  &  &  25.37\\[0.35ex]
    GPT2-xl-uniform-2 &  & $\{0, 10, 20, 30, 40, 46\}$ &  &  &  &  &  26.07  \\[0.35ex]
    GPT2-xl-6bh &  & $\{0, 4, 9, 13, 18, 23\}$ &  &  &  &  &  25.54 \\[0.35ex]
    GPT2-xl-6th &  & $\{24, 28, 33, 37, 42, 47\}$ &  &  &  & & 47.51\\[0.35ex]
    GPT2-xl-random &  & $\{2, 11, 17, 25, 34, 40\}$ &  &  &  &  & 42.48 \\[0.35ex]
    \midrule
    GPT2-large-psudo &\multirow{6}{*}{6} & $\{0, 6, 12, 23, 29, 35\}$ & \multirow{6}{*}{20}& \multirow{6}{*}{1,280} & \multirow{6}{*}{183 M} &\multirow{6}{*}{6} &  \bf 23.79\\[0.35ex]
    GPT2-large-uniform &  & $\{0, 7, 14, 21, 28, 35\}$ &  &  &  &  &  27.68\\[0.35ex]
    GPT2-large-uniform-2 &  & $\{0, 8, 16, 24, 32, 34\}$ &  &  &  &  &  27.78\\[0.35ex]
    GPT2-large-6bh &  & $\{0, 2, 4, 7, 9, 12\}$ &  &  &  &  &  35.88\\[0.35ex]
    GPT2-large-6th &  & $\{13, 17, 21, 26, 30, 35\}$ &  &  &  &  &  74.15\\[0.35ex]
    GPT2-large-random &  & $\{5, 25, 26, 29, 32, 34\}$ &  &  &  &  & 79.68\\[0.35ex]
    \midrule
    GPT2-medium-psudo &\multirow{6}{*}{6} & $\{0, 4, 8, 15, 19, 23\}$ & \multirow{6}{*}{16}& \multirow{6}{*}{1,024} & \multirow{6}{*}{128 M} & \multirow{6}{*}{6} & \bf 28.09 \\[0.35ex]
    GPT2-medium-uniform &  & $\{0, 4, 8, 12, 16, 20\}$ &  &  &  &  &  35.83 \\[0.35ex]
    GPT2-medium-uniform-2 &  & $\{0, 5, 10, 15, 20, 23\}$ &  &  &  &  & 35.4 \\[0.35ex]
    GPT2-medium-6bh &  & $\{0, 2, 4, 6, 8, 10\}$ &  &  &  &  & 36.2 \\[0.35ex]
    GPT2-medium-6th &  & $\{12, 14, 16, 18, 20, 23\}$ &  &  &  &  & 88.71 \\[0.35ex]
    GPT2-medium-random &  & $\{1, 3, 7, 11, 15, 21\}$ &  &  &  &  & 75.13 \\[0.35ex]
    \midrule
    GPT2-psudo &\multirow{4}{*}{6} & $\{0, 2, 4, 7, 9, 11\}$ & \multirow{4}{*}{12}& \multirow{4}{*}{768} & \multirow{4}{*}{81 M} &  \multirow{4}{*}{6} & \bf 26.14\\[0.35ex]
    GPT2-uniform &  & $\{0, 2, 4, 6, 8, 10\}$ &  &  &  &  & 26.6 \\[0.35ex]
    GPT2-6bh &  & $\{0, 1, 2, 3, 4, 5\}$ &  &  &  &  &  29.86 \\[0.35ex]
    GPT2-6th &  & $\{6, 7, 8, 9, 10, 11\}$ &  &  &  &  &  49.61\\
    \bottomrule
\end{tabular}
\end{center}
\label{tab:trun-finetune}
\end{table}
We list several observations from fine-tuning results:
\begin{description}
\item [The best validation curves come from the "uniform/bottom-half/pseudo-uniform" strategies] We observe a better convergence in these settings, which is similar to previous paper reported results.
\item [The "pseudo-uniform" strategy achieves the best test results after fine-tuning] Perplexities are the lowest in this setting, as shown in table \ref{tab:trun-finetune}.
\item [The "bottom-half" strategy outperforms the "top-half"] According to previous report, this is due to the fact that bottom layers tend to learn embeddings and general representations while top layers are more task-specialized.
\end{description}
Overall, as a conclusion of this experiment the pseudo-uniform initialization scheme clearly allows for better generalization (table \ref{tab:trun-finetune}). We can also conclude that pre-training plays a significant role in aligning the weights and making convergence faster: the performance of DistilGPT2 supports this claim.

\end{document}